# Geometric Artifact Correction for Symmetric Multi-Linear Trajectory CT: Theory, Method, and Generalization

Zhisheng Wang[a,b], Yanxu Sun[a,b], Shangyu Li[a,b], Legeng Lin[a,b], Shunli Wang[a,b, *], and Junning Cui[a,b,*]

a. *Center of Ultra-precision Optoelectronic Instrument engineering, Harbin Institute of Technology, Harbin, 150080, China*
b. *Key Lab of Ultra-precision Intelligent Instrumentation, Harbin Institute of Technology, Harbin, 150080, China*

*Abstract*—**For extending CT field-of-view to perform non-destructive testing, the Symmetric Multi-Linear trajectory Computed Tomography (SMLCT) has been developed as a successful example of non-standard CT scanning modes. However, inevitable geometric errors can cause severe artifacts in the reconstructed images. The existing calibration method for SMLCT is both crude and inefficient. It involves reconstructing hundreds of images by exhaustively substituting each potential error, and then manually identifying the images with the fewest geometric artifacts to estimate the final geometric errors for calibration. In this paper, we comprehensively and efficiently address the challenging geometric artifacts in SMLCT, , and the corresponding works mainly involve theory, method, and generalization. In particular, after identifying sensitive parameters and conducting some theory analysis of geometric artifacts, we summarize several key properties between sensitive geometric parameters and artifact characteristics. Then, we further construct mathematical relationships that relate sensitive geometric errors to the pixel offsets of reconstruction images with artifact characteristics. To accurately extract pixel bias, we innovatively adapt the Generalized Cross-Correlation with Phase Transform (GCC-PHAT) algorithm, commonly used in sound processing, for our image registration task for each paired symmetric LCT. This adaptation leads to the design of a highly efficient rigid translation registration method. Simulation and physical experiments have validated the excellent performance of this work. Additionally, our results demonstrate significant generalization to common rotated CT and a variant of SMLCT.**

## I. Introduction

X-ray industrial computed tomography (CT) is an advanced non-destructive testing (NDT) technique used to intuitively and clearly reveal an object's internal structure. It is widely applied in industries such as precision manufacturing [1], aerospace [2], automotive [3], nuclear [4], electronics [5], and so on. Due to this competitive advantage, many multimodal and non-standard scanning CTs have been profoundly and widely developed for different requirements until now [6]. As a typical successful representative in the non-standard CT scanning modes, Multi-Linear trajectory CT (MLCT) has a significant advantage and easy implementation to enlarge the field-of-view (FOV) for NDT, as displayed in Fig. 1(a)-(c) [7–9]. In theory, utilizing the same property as the second-generation CT, MLCT can conveniently extend the FOV over a large range, making it suitable for measuring various sized and shaped objects [10–12]. What is even more advantageous is that, unlike second-generation CT, the detector of MLCT remains stationary during each LCT scan, which eliminates the need to rebin multiple sets of truncated projection, avoiding more errors [8,13]. To achieve ultra-large extended FOV imaging beyond that of conventional CT scanning modes (e.g., FOV magnification exceeding twice, see Fig. 1(d)-(g)), as a special case of MLCT, the full-scan Symmetric MLCT (SMLCT) geometry, along with its differentiation-based filtered backprojection reconstruction, was designed and validated to yield optimal results [10]. Therefore, this paper focuses on the SMLCT scanning mode as the research subject. However, in practical imaging, a prerequisite for the normal operation of this work is precise geometric correction; otherwise, severe geometric artifacts will make it difficult for us to see details clearly, as shown in Fig. 1(g).

Many methods have been developed for CT geometric artifact correction, which can be divided into two categories according to whether they use phantoms for correction: phantom-based correction methods [14–19] and non-phantom self-correction methods [20–24]. Phantom-based methods can accurately calibrate some geometric parameters but require additional scanning and dedicated phantoms. Compared with some phantom-based correction methods and even precise mechanical systems or optical or electromagnetic tracking systems, non-phantom methods do not require a dedicated phantom or any other specific markers inside the measured object and are low-cost solutions that are advantageous for practical applications [25]. For example, in the common rotated CT (RCT), the most sensitive geometric parameter—the position of projection center of rotation (COR)—can lead to some ghost artifacts and blurr high-frequency details in the reconstruction image [22]. Some main efficient correction methods include the center-of-sinogram method, geometry method, opposite-angle method, and projection image cross-correlation registration method [22,26–31]. However, these methods are only applicable to conventional RCT modes and make it difficult to directly introduce our special SMLCT geometry. Moreover, these approaches are generally used to correct the geometric deviation of COR, where we can see from the actual SMLCT reconstructed slice that there are other severe geometric artifact features beyond ghost artifacts (see Fig. 1(g)) [32]. Recent approaches for CT geometric correction involve iterative algorithms [19,24,33,34] and deep learning models [35,36]. Iterative methods optimize the projection-data consistency, while deep learning predicts correction parameters directly from raw data. However, their practical application in engineering is limited due to high computational cost and the need for extensive training data.

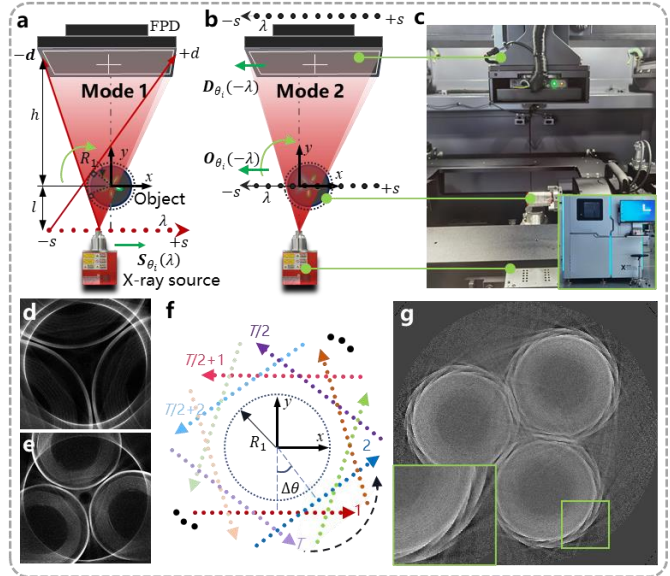

Fig. 1. Reconstruction with different CT scanning modes. (a) and (b) Schematics of source (**Mode 1**) and object-FPD (**Mode 2**) linearly scanning in MLCT, (c) on-the-shelf CT device with multiple degrees of freedom in motion, (d) and (e) are reconstructed images via the RCT and FPD-biased RCT scanning and geometric calibration, (f) and (g) source trajectory and reconstructed image of SMLCT.

In industrial applications of MLCT for large objects, ensuring precise motion between two trajectory segments is challenging for mechanical and control systems. To address this, Li *et al.* proposed a self-calibration method that uses invariant moments of the attenuation distribution image to measure motion. The accuracy of this method improves with increased projection data, but it only corrects the rotation angle [25]. Yu *et al.* introduced a simple yet labor-intensive approach, where geometric deviations are sequentially substituted within a certain range for reconstruction [13]. Each updated geometric error is labeled in a new reconstructed image. By subjectively evaluating hundreds of these reconstructed images, the optimal set of geometric errors is determined. Current geometric correction theories and methods primarily target standard CT scanning trajectories, with a lack of theoretical and efficient approaches for geometric correction in linear scanning trajectory CT. Thus, the motivation of this paper is to design an automatic, accurate, and efficient non-phantom geometric artifact correction method for SMLCT, serving engineering practice well.

According to the theory of visible and invisible boundaries [37], we notice that symmetric LCT reconstruction images contain a large amount of redundant data [9,10]. Leveraging or registering redundant data that carries identical geometric information appears to be a key breakthrough in developing an effective non-phantom geometric correction method for SMLCT. Most image cross-correlation registration methods for RCT geometric correction use primarily iteratively aligned forward projection and meausred projection images to identify the geometric errors corresponding to the maximum similarity [20,24,31,38,39]. For SMLCT, a different and interesting insight is that inverse geometric errors can be traced back to the reconstruction images, specifically the limited-angle scanned LCT images in SMLCT. This insight involves precisely registering the two symmetric LCT images to estimate the pixel offsets between them, and then constructing mathematical relationships to correlate these offsets with the geometric errors requiring correction. Moreover, new questions arise, such as which geometric errors need to be estimated through image registration and how to achieve more accurate image registration.

In this paper, some main contributions are summarized as:

1) We analyze and summarize the formation mechanisms and key properties of geometric artifacts by investigating the distribution characteristics of reconstructed data in MLCT with various geometric errors. This analysis provides essential theoretical guidance for geometric error inversion in MLCT-type scanning modes.

2) Assuming redundancy between reconstructed images and their accurate registration, we construct mathematical relationships that link pixel offsets between symmetric LCT reconstruction images to sensitive geometric errors. This enables the inversion of geometric errors in the image domain.

3) We introduce the Generalized Cross-Correlation with Phase Transform (GCC-PHAT) algorithm, originally developed for time-lag estimation in sound processing, into our rigid translation image registration task. By leveraging the known valid Fourier data regions of LCT, this approach effectively weights the cross-power spectrum between two symmetrical LCT reconstruction images, enhancing the sharpness of the cross-correlation surface (CCS) and yielding precise pixel offsets.

4) Our work can also be broadly generalized to other CT scanning modes, such as common RCT—including full- and half-scan—and variants of SMLCT.

The rest of this paper is organized as follows. Section 2 briefly describes the special SMLCT geometry in the MLCT scanning modes and its reconstruction for extending large FOVs. In Section 3, we detail our geometric artifact calibration theories and methods. Section 4 conducts some simulated and real-data experiments and further generalizes our method to other CT scanning modes. Finally, Sections 5 and 6 present the discussion and conclusion, respectively.

## II. Preliminaries

### A. MLCT scanning modes

On an off-the-shelf industrial CT equipped with a micro-focal X-ray source and multiple degrees of freedom in motion (see Fig. 1(c)), there are two simple implementation modes for MLCT that conveniently extend the FOV over a large range.

**Mode 1.** The source translation trajectory $S_{\theta_i}(\lambda)$ in Fig. 1(a) is formulated as:
$$S_{\theta_i}(\lambda) = [\lambda, -l, 0] \cdot R_{\theta_i}, \quad \lambda \in [-s, s]. \tag{1}$$

**Mode 2.** It is also equivalent synchronous motions of the flat panel detector (FPD) $D_{\theta_i}(-\lambda)$ and object $O_{\theta_i}(-\lambda)$ along the reverse direction of the above source translation, shown in Fig. 1(b) [10],
$$D_{\theta_i}(-\lambda) = [-\lambda, h, 0] \cdot R_{\theta_i}, \quad O_{\theta_i}(-\lambda) = [-\lambda, 0, 0] \cdot R_{\theta_i}, \tag{2}$$

where $\lambda$ is translation coordinate, $R_{\theta_i}$ is the rotation matrix and $\theta_i$ denotes the angle between the x-axis and translation trajectory, $i = 1, 2, \ldots, T$. The FOV radius with sufficient data is determined as [7]
$$R_1 = \frac{sh - dl}{\sqrt{(l+h)^2 + (s+d)^2}}, \tag{3}$$

where $s$ and $d$ are the half-length of the source (or object-FPD) linear trajectory and FPD, and $l$ and $h$ are the verical distances from the iso-center $o$ to the source and FPD, respectively.

### B. SMLCT geometry and reconstruction for extending large FOVs

To achieve optimal 3D imaging, the SMLCT geometry with the smooth redundancy-weighted differentiation-based filtered backprojection reconstruction needs to be applied [10],
$$f(\vec{x}) = \sum_{i=1}^{T} \frac{1}{4\pi L} \int_{-s}^{+s} \int_{-d}^{+d} w_g(l, h, \lambda, u, v) \cdot \frac{\partial}{\partial u}\left(p_{\theta_i}(\lambda, u, v) \cdot \mathcal{R}_{\theta_i}(l, h, \lambda, u)\right) \cdot \hbar_H(u' - u) du d\lambda, \tag{4}$$

where $w_g(\cdot)$ is the geometry weight, and $\mathcal{R}_{\theta_i}(\cdot)$ is the redundancy weighting function, some details can be found in Refs. [9,11]. $L = -x\sin\theta + y\cos\theta + l$. $\hbar_H(\cdot)$ is Hilbert filtering kernel. To form the SMLCT geometry, $T$ can be determined as:
$$T = \text{ceil}\left(\frac{2\pi}{\Delta\theta}\right) + T_e, \quad \Delta\theta = 2\arctan\left(\frac{d}{h}\right). \tag{5}$$

where $T_e$ is adjusted to generate sufficient redundant data for weakening truncation errors and producing an even number [10].

When scanning bundled lithium batteries using three different modes on an on-the-shelf CT system, both the standard RCT (Fig. 1(d)) and the FPD-biased RCT (Fig. 1(e), with an FOV magnification of less than two), fail to fully cover the objects, resulting in severe cupping artifacts after reconstruction. Although SMLCT achieves over 3× FOV magnification to cover the entire object, as shown in Fig. 1(f) and (g), it reveals severe and complex artifacts (i.e., more than one sensitive error term) when reconstructed using ideal geometric parameters.

## III. Theories and Methods

### A. Identifying sensitive geometric error terms

Figure 2 depicts nine terms of geometric errors in LCT. It is difficult and undesirable to calibrate so many errors in engineering. A commonly used simplification method in applications is to extract a few sensitive geometric error terms to eliminate the geometric artifacts [40,41]. Therefore, we identify the most sensitive error terms through experiments controlling variables. Figure 3 shows reconstructed horizontal slices for the Shepp-Logan phantom, metrics, including the root mean square error (RMSE) and structural similarity index (SSIM), and key border profiles. Comparing these results, notice that the most sensitive two terms, i.e., $\delta l$ and $\delta s$, lead to severe windmill-shaped and ghost artifacts, respectively, followed by $\delta u$ (also yields ghost), $\theta_\lambda$, and $\delta h$, and finally $\theta_d$, $\delta v$, $\theta_{out}$, and $\theta_{in}$.

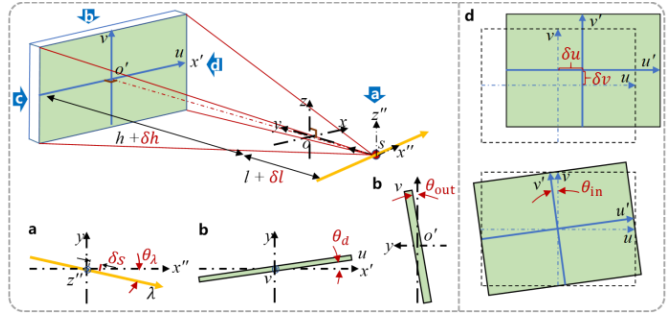

Fig. 2. Geometric errors of LCT. $\delta l$ and $\delta h$ are key error terms influencing the magnification ratio; $\delta s$ and $\theta_\lambda$ are errors of the linear trajectory; and $\delta u$, $\delta v$, $\theta_{in}$, $\theta_{out}$, and $\theta_d$ are error terms of FPD.

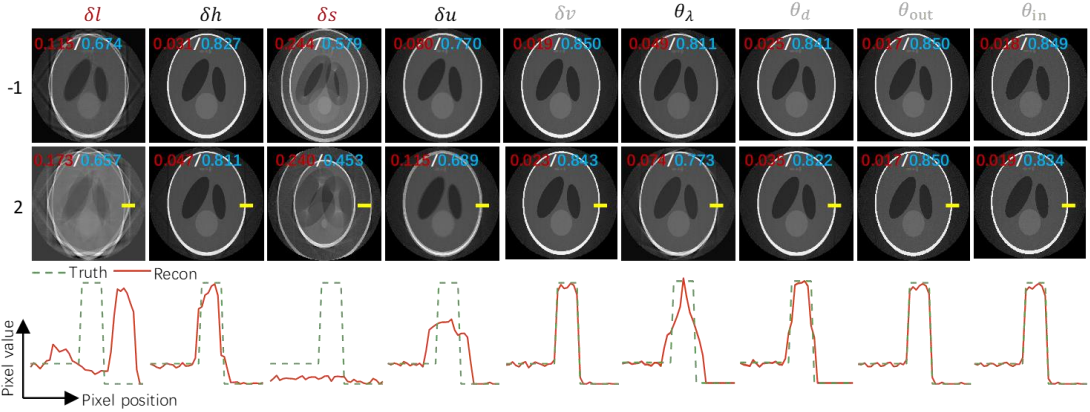

Fig. 3. Reconstruction results with different geometric error terms and values. For $\delta l$, $\delta h$, $\delta s$, $\delta u$, and $\delta v$, we repectively set -1 mm and 2 mm. For $\theta_\lambda$, $\theta_{in}$, $\theta_{out}$, and $\theta_d$, the unit of these two values is dregree. For the metric in each slice, the red and blue values are RMSE and SSIM. The 3rd row exhibits the profiles marked in the reconstructed images located in the 2nd row.

*B. Formation mechanisms of geometric artifacts in MLCT*

To guide our correction efforts effectively, it is essential to analyze the formation mechanisms of the windmill-shaped and ghosting artifacts caused by the aforementioned sensitive geometric error terms. It is known that LCT is a limited-angle scan (see Fig. 4(a)), and the relationship between its projection data and the reconstructed image satisfies the following "visible and invisible boundary" theorem:

**Theorem 1.** *In limited-angle scanned CT reconstructed images using analytical algorithms, the object boundaries along the direction tangent to the available rays are reconstructable, whereas the image boundaries along the direction tangent to the missing rays remain invisible* [37].

Based on **Theorem 1**, Figure 4(a) illustrates the visible boundaries and fan-type regions (these regions are equivalent to the 2D Radon space in a sense [42]) of the 1st LCT, while Fig. 4(b) highlights the missing data regions in the Fourier domain of the reconstructed image from Fig. 4(a). In Fig. 4(a), the two angles of visible boundaries relative to the central virtual detector can be determined based on the geometric relationship,

$$\alpha = \arctan(\frac{d^2 - R_1^2}{dh + R_1\sqrt{d^2 + h^2 - R_1^2}}), \quad \vartheta = \frac{\pi}{2} - \arctan\left(\frac{l+h}{s+d}\right). \tag{6}$$

It is observed that Fig. 4(c) displays numerous overlapping (or redundant) regions of LCT from the SMCT geometry, particularly between any two symmetric LCTs. This observation is a key point for indirectly correcting geometric artifacts by registering the two reconstructed images.

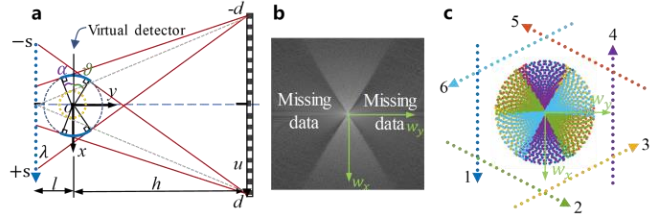

Fig. 4. Description of visible boundaries, regions, and Fourier data in LCT and MLCT. (a) depicts visible boundaries tangent by rays and fan-type Radon regions in LCT, where the dashed line is a circular FOV, and the solid line above it is the visible boundary; (b) Missing data in the 2D Fourier space of LCT; and (c) shows some linear scanning trajectories and the distribution of fan-type Radon regions in SMLCT.

**Formation mechanisms of windmill-shaped artifacts**: When the distance from the iso-center $o$ to the ideal source linear trajectory exceeds that to the real trajectory by a positive bias $\delta l$ and reconstruction is performed based on the ideal geometric positional mapping (e.g., $u'$ is relative to $s'$), a positive shift $\rho$ of the reconstructed iso-center $o'$ relative to the ideal iso-center occurs along the y-axis (or the central ray direction), as illustrated in Fig. 5(a). By analogy, with this distance bias of the source-to-iso-center being negative (i.e., $-\delta l$), there will be a negative shift $-\rho$ of $o'$ compared to $o$. Additionally, we can find that the iso-center bias produced by $\delta h$ is similar to that of $\delta l$, and due to the geometric ratio between the source and FPD, this results in a weaker bias compared to $\delta l$ at the same values (Fig. 5(b)). Due to the existing geometric errors influencing the magnification ($\delta l$ and $\delta h$), as shown in Fig. 5(c) and (d), each reconstructed iso-center $o'$ of LCT will be distributed around a circle radius with $\rho$ in SMCLT reconstruction. In this way, Fig. 5(c) clearly illustrates the misalignment of the fan-type regions reconstructed from each segment LCT of MLCT. This misalignment indicates that the visible boundaries of each LCT do not transition or connect properly, resulting in windmill-shaped artifacts.

**Formation mechanisms of ghosting artifacts**: When the real source trajectory is offset along the positive motion relative to its ideal scan, producing an error $\delta s$. If reconstruction still adheres to the ideal geometric relationship, a pixel offset (or iso-center $o'$ offset) $\alpha$ along the positive x-axis (the linear scanning direction) emerges in the real reconstructed image as compared to the ideal one (Fig. 5(e)). If we let the detector make a positive bias $\delta u$, the real reconstructed image also has a pixel offset, and this offset degree is weaker than the one of $\delta s$ at the same volume (Fig. 5(f)). At the virtual detector, reconstruction images from two symmetric LCTs (e.g., 1st and $(T/2+1)^{th}$ LCT) with the source trajectory bias will appear an iso-center $o'$ offset of $2\alpha$ (Fig. 5(g)), resulting in ghosting artifacts similar to those seen in RCT.

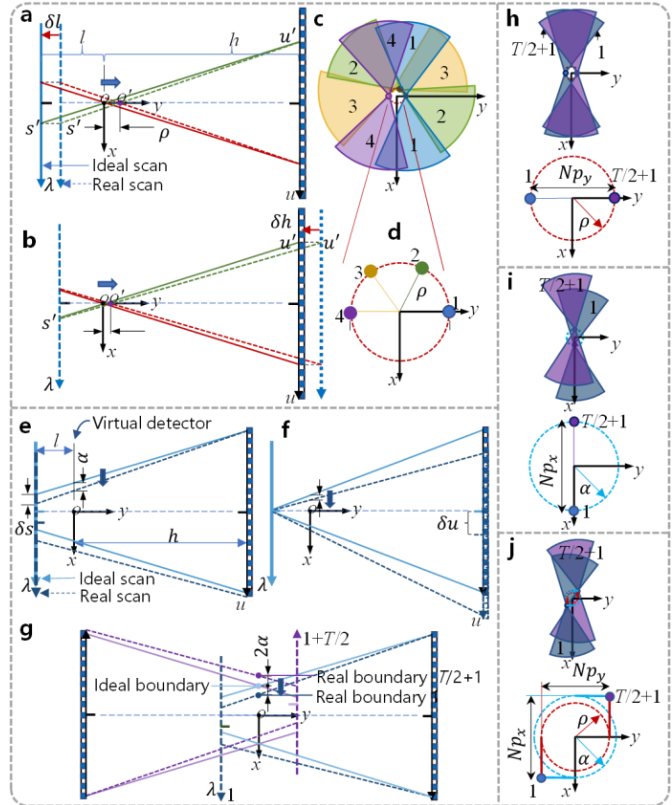

Fig. 5. Description of the formation mechanisms for windmill-shaped and ghosting artifacts in MLCT. (a) and (b) Schematics of iso-center $o'$ bias along the central ray due to $\delta l$ and $\delta h$, respectively; (c) and (d) are map of fan-type Radon regions and magnified distribution map of iso-center $o'$ in MLCT due to the existence of $\delta l$ and $\delta h$, respectively; (e) and (f) are schematics of the pixel offsets with $\delta s$ and $\delta u$, respectively; (g) Illustration of the pixel offsets $2\alpha$ of two symmetric LCT on the virtual detector locations with $\delta s$; (h), (i), and (j) are schematics of iso-center offset due to $\delta l$ and $\delta h$, $\delta s$ and $\delta u$, and comprehensive above all error terms, respectively. In (h)-(j), fan-type Radon regions and their iso-center locations of two symmeric LCTs are drawn in each figure.

Based on the above analysis of the formation mechanisms for the two types of artifacts, we summarize the following key properties of the general MLCT geometry:

**Property 1.** *Two geometric errors, $\delta l$ and $\delta h$, affecting the geometric magnification can cause the reconstructed iso-center $o'$ to deviate from the ideal fixed coordinate origin $o$ along the direction perpendicular to the central ray of each linear trajectory. As a result, some reconstructed iso-center $o'$ in MLCT will be distributed along an error circle with a radius $\rho$ (see Fig. 5(d)).*

**Property 2.** *Geometric errors $\delta s$ and $\delta u$ can cause the reconstructed iso-center $o'$ to shift along directions parallel to their respective linear trajectories. This shift is orthogonal and uncorrelated with the offset described in **Property 1** (see Fig. 5(h)–(j)). As a result, some reconstructed iso-center $o'$ will be distributed along an error circle with a radius $\alpha$ in MLCT.*

**Property 3.** *Regarding geometric errors $\delta l$ and $\delta h$, the precision of geometric magnification is crucial for preventing the iso-center bias described in **Property 1**. Therefore, selecting either $\delta l$ or $\delta h$ for correction to achieve accurate geometric* *magnification can simplify the overall correction process. Similarly, selecting either $\delta s$ or $\delta u$ for correction can avoid the iso-center bias described in* **Property 2**.

## C. Sensitive geometric errors inverse in SMLCT

According to **Properties 1–3**, two sensitive geometric errors $\delta l$ and $\delta s$ can independently make the iso-center $o'$ bias in the two orthogonal directions. Therefore, by estimating pixel offsets through image registration and establishing mathematical relationships between the iso-center offsets and geometric errors, geometric error inversion can be achieved.

Figure 5(h)-(j) respectively depicts the distribution of two iso-centers of symmetric LCTs with a vertical linear trajectory under three scenarios: carrying only the errors $\delta l$ and $\delta h$, carrying only the errors $\delta s$ and $\delta u$, and carrying all of the above errors combined. Cleverly, if the pixel offsets $(Np_y, Np_x)$ (unit: pixels) are known, the radii of the two error circles, $\rho$ and $\alpha$, can be easily and directly obtained, enabling the construction of geometric error inversion relationships in SMLCT.

Regarding the relationship between $\delta l$ and $\rho$, given the number $Np_y$ of horizontal pixel offsets between the iso-centers of two symmetric LCTs (see the red circle diameter in Fig. 5(h)), we can derive this relationship based on the geometries shown in Fig. 5(a) – (d),

$$\frac{h-\rho}{l+\delta l+\rho} = \frac{u'}{s'} = \frac{h}{l}, \quad \text{with} \quad \rho = \frac{\Delta y \cdot Np_y}{2}, \tag{7}$$

where $\Delta y$ indicates the pixel width, and it can be determined by $\Delta y = 2R_1/N$, where $N$ is the image width. The intermediate variables $u'$ and $s'$ imply the original geometric relationship to keep the reconstruction iso-center $o'$ overlap with the origin $o$ (i.e., assuming no this bias). Thus, equation (7) can be simplified as

$$\delta l = -\frac{(l+h) \cdot R_1 \cdot Np_y}{hN}. \tag{8}$$

Regarding the mathematical model relating $\delta s$ and $\alpha$, if the number $Np_x$ of vertical pixel offsets is known (indicated by the diameter of the blue circle in Fig. 5(i)), it can be expressed based on the geometric relationships shown in Fig. 5(e)–(g),

$$\frac{\alpha}{\delta s} = \frac{h}{l+h}, \quad \text{with} \quad \alpha = \frac{\Delta x \cdot Np_x}{2}, \tag{9}$$

where $\Delta x$ is the pixel height, meeting $\Delta x = 2R_1/M$, $M$ is the image height. The final form is expressed as

$$\delta s = \frac{(l+h) \cdot R_1 \cdot Np_x}{hM}. \tag{10}$$

Based on the premises that calibrated geometric errors are usually considered fixed and rotation accuracy is high (i.e., angle error is ignored) during scanning, two important properties are summarized as follows:

**Property 4.** *For paired images of tilted symmetric LCTs (or the non-1st paired symmetric LCTs), if they are rotated to align with the paired reconstructed images of the 1st symmetric LCTs, theoretically, the estimated geometric errors should be equal.*

**Property 5.** *The reconstruction images of symmetric LCTs can achieve feature alignment through simple rigid translation without the need for complex transformations (e.g., deformation or rotation). This allows pixel offsets to be calculated using rigid image translation registration.*

In fact, **Property 4** indicates that we can utilize all symmetrical LCTs to calculate geometric errors and improve estimation accuracy, thus proposing the following technique:

**Technique 1.** *Rotate the reconstructed images of all non-1st paired symmetric LCTs by an angle opposite to their original scan to align them with the image orientation of the 1st paired symmetric LCTs (i.e., the 1st and (T/2+1)th LCTs). Then, apply Eqs. (8) and (10) to estimate sets of geometric errors. A criterion is designed to select the optimized final result:*

**Criterion 1.** *For $Np_x$ or $Np_y$, if the number of equal values is greater than or equal to ceil(T/4), it is set as the final value; Otherwise, the mean is taken.*

## D. GCC-PHAT-based image registration for symmetric LCTs

To compute geometric errors $\delta l$ and $\delta s$, we need to exactly obtain the number of pixel offsets $(Np_x, Np_y)$ by registering two reconstruction images of symmetric LCTs. The normalized cross-correlation (NCC, and CC is the cross-correlation) is a classic statistical registration algorithm that determines the degree of matching by calculating CCS, $R_{fg}(m,n)$) between a fixed image $f(m,n)$ and a moving image $g(m,n)$, thereby being suitable for our rigid translation registration task [24,27,38]. The horizontal and vertical coordinates corresponding to the maximum in $R_{fg}(m,n)$ (i.e., the highest peak) are the required pixel offset numbers,

$$(Np_x, Np_y) = \underset{m,n}{\operatorname{argmax}}(R_{fg}(m,n)). \tag{11}$$

Nevertheless, for actual noisy images, the accurate highest peak of CCS derived from NCC is easily affected, manifested as lacking sharpness and masked by surrounding enhanced pseudo-peaks, leading to inaccurate estimation. Figure 6 exhibits some results for some symmetric LCT images, including the simulated noisy Shepp-Logan phantom and real flower bud. These programs were achieved on the Astra toolbox [39] and run on a PC equipped with the 12th Gen Intel (R) Core™ i9-12900KF CPU at 3.20 GHz and the NVIDIA GeForce RTX 3090 Ti GPU, and their running times are listed. In Fig. 6(a) and (b), NCC has high efficiency, but it cannot achieve great registration in noisy simulated and real images. Moreover, intensity-based iterative registration (IIR) is also a common method, but selecting appropriate initial parameters is generally challenging, making it difficult for the loss function to converge, and the iterative process is time-consuming [43]. Figure 6(c) shows the failed registration result of IIR and long running time (Optimizer: regular step gradient descent, Initial radius: $5.1 \times 10^{-4}$, Iterations: 600, Metric: mean squares).

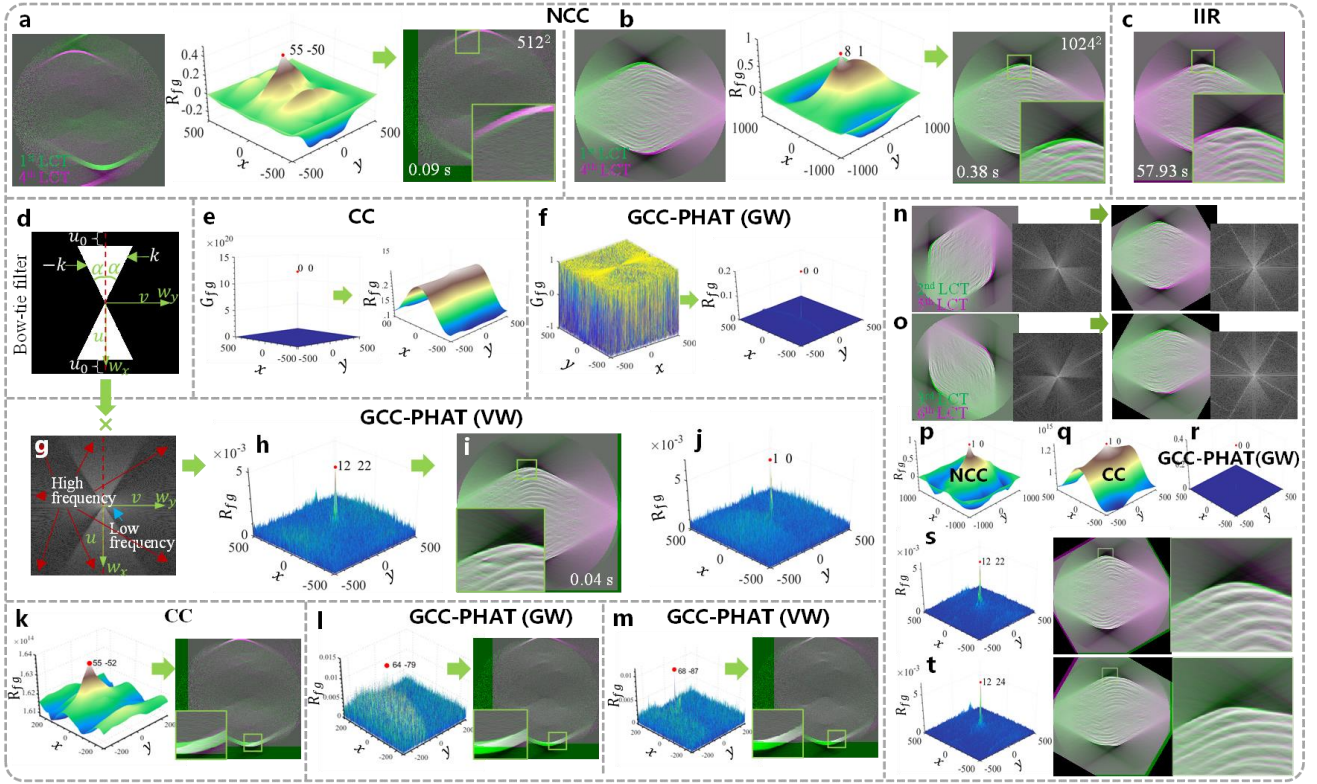

Fig. 6. Registration results of different methods. (a) and (b) are the NCC registration results for the 1st paired LCT reconstructed noisy Shepp-Logan phantom and real flower bud data, respectively; (c) Registration result of IIR for the 1st paired LCT reconstructed real flower bud data; (d) Bow-tie filter in the frequency spectrum; (e) and (f) maps of the cross-power spectrum and CCS using CC and GCC-PHAT (GW) for real flower bud data, respectively; (g) – (i) Cross-power spectrum, CCS, and final registration result of the 1st paired LCT bud images using GCC-PHAT (VW) with the described bow-tie filter in (d); (j) CCS map calculated using GCC-PHAT (VW) with a bow-tie filter and its upper and lower boundary regions not zeroed, (k) – (m) are the CCS maps and registration results of the 1st paired LCT simulated noisy Shepp-Logan phantom via CC, GCC-PHAT (GW), and GCC-PHAT (VW), respectively; (n) and (o) non-1st (i.e., the 2nd and 3rd) paired LCT images and corresponding inverse rotated images via Technique 1, including the overlaying image and cross-power spectrum; (p) – (r) are calculated CCS maps of (n) related to NCC, CC, and GCC-PHAT (GW), respectively; (s) and (t) are CCS maps and registration results using GCC-PHAT (VW) for (n) and (o), respectively.

To address it, we draw inspiration from the sound positioning and speech processing fields, which utilized the basic idea of GCC-PHAT to improve time-delay estimation accuracy in noisy conditions [44–46]. To clarify it, we define two new nouns:

**VFR**: *The areas where the main frequency components of the two signals overlap or are most similar in the cross-power spectrum, i.e., valid frequency regions.*

**IFR**: *The spectrum regions containing frequency components from the image background and noise, i.e., invalid frequency regions.*

PHAT means a weighting (or whitening) to normalize the magnitude components of the cross-power spectrum. Based on the premise that the speech frequency range is known, this **VFR** is weighted, sharpening (strengthening) the CCS peak and reducing sensitivity to noise. Due to the elusive frequency components in images, identifying the required **VFR** while avoiding **IFR** is generally challenging.

We notice that the core idea of GCC-PAHT can be applied to better match images for limited-angle scanned CTs like SMLCT. In LCT, **VFR** can be maximally deciphered from the Fourier domain of reconstruction images according to **Theorem 1**, as shown in Fig. 4(a) and (b).

In 2D Fourier space, we construct a bow-tie filter $B(u,v)$ (a bow-tie-shape window function, as displayed in Fig. 6(d)) to multiply by the cross-power spectrum $G_{fg}(u,v)$ ($u$ and $v$ are discrete frequency points relative to the coordinates $w_x$ and $w_y$). Its boundaries are defined by different lines, and their angle can be determined by referring to Fig. 4(a) and Eq. (6). We use the intersection areas of two images, which are limited to a bow-tie-shape region with an angle $2\alpha$, as drawn in Fig. 6(d). To improve estimation accuracy and robustness, the design of GCC-PHAT should consider a criterion:

**Criterion 2**. *In GCC-PHAT, the whitened areas in the cross-power spectrum are supposed to include more* **VFR** *(i.e.,* **VFR** *is weighted, VW) rather than the global frequency band (i.e., globally weighting, GW) or introduce more* **IFR**; *otherwise, the noise power may be stronger than* **VFR** *after whitening, leading to inaccurate estimation.*

Based on **Criterion 2**, we design a GCC-PAHT (VW) formula,

$$R_{fg}(m,n) = \mathcal{DFT}^{-1}\left\{\frac{G_{fg}(u,v)}{|G_{fg}(u,v)|} \times B(u,v)\right\}, \tag{12}$$

where $G_{fg}(u,v) = \mathcal{DFT}(f(m,n)) \cdot \overline{\mathcal{DFT}(g(m,n))}$, which describes the correlation of two signals in the Fourier domain. $\mathcal{DFT}(\cdot)$ and $\mathcal{DFT}^{-1}(\cdot)$ are the discrete Fourier transform and its inversion. $1/|G_{fg}|$ is the PHAT weighting, essentially a whitening filter. Without this term, equation (12) is just a CC function. One implementation form of $B(u,v)$ is

$$B_1(u,v) = \begin{cases} 1, & v < u/k \\ 0, & v < u/k \end{cases}, B_2(u,v) = \begin{cases} 0, & v < u/-k \\ 1, & v > u/-k \end{cases},$$
$$k = \tan\left(\frac{\pi}{2} - \alpha\right), u \in [-u_0, U - u_0], \tag{13}$$

where $B(u,v) = B_1(u,v) \odot B_2(u,v)$, and $\odot$ denotes the logical same or operation. According to **Criterion 2**, $u_0$ defines the width of the zeroing boundary at the top and bottom of $B(u,v)$, as described in **Technique 2** below.

**Technique 2**. *With a low signal-noise-ratio, a zeroing region with a width of $u_0$ (see Eq. (13) and Fig. 6(d)) is set at the top and bottom of $B(u,v)$ to prevent the introduction of excessive high-frequency noise at the boundaries during whitening. Generally, $u_0$ can be set to 1/8 of the image width.*

To match the 1st paired symmetric LCT reconstructed bud images, the CC peak is smooth and difficult to extract, as shown in Fig. 6(e), and the estimated result of GCC-PHAT (GW) is invalid (0, 0) in Fig. 6(f). Because these edged areas usually include the high-frequency noise and information (see Fig. 6(g)). To reduce the interference from high-frequency noise from edge regions, we get accurate pixel offset estimation and the final registration results through **Technique 2**, as depicted in Fig. 6(h) and (i), respectively. Without **Technique 2**, the intensity of **IFR** would be whitened to nearly match that of the mid-range low-frequency **VFR**, leading to incorrect pixel offset estimation from the CCS highest peak (e.g., see (1, 0) in Fig. 6(j)). Furthermore, we assess the performance of CC and GCC-PHAT (GW and VW) for the noisy simulation phantom; their corresponding CCS figure and registration result, are shown in Fig. 6(k), (l), and (m), respectively. Figure 6(m) demonstrates the best registration effect, which further strengthens the effectiveness of GCC-PHAT (VW).

Based on **Property 4** and **Technique 1**, for the other paired reconstructed images of non-1st symmetric LCTs (Fig. 6(n)), we observe that NCC, CC, and GCC-PHAT (GW) all perform poorly (as shown in Fig. 6(p), (q), and (r), respectively). In contrast, GCC-PHAT (VW) effectively avoids the inaccuracies caused by image rotation transformation and extrapolation (see Fig. 6(s) and (t)).

*E. Implementation framework of our correction method*

Figure 7 illustrates the implementation framework of the proposed geometric artifact correction method for SMLCT. By registering the reconstruction images of each paired LCT to compute the pixel offsets, we can infer the sensitive geometric errors and use them to correct the geometric artifacts. The main steps are as follows.

1) For all paired reconstruction images of symmetric LCTs using ideal geometric parameters, perform individually the inverse rotation operation $\mathcal{R}ot_{\theta_j}^{-1}(\cdot)$ on each paired image with the angle $\theta_j = (j-1) \cdot \mathcal{R}_{dir} \cdot \Delta\theta, j \in [1, T/2]$. Here, $\mathcal{R}_{dir}$ is a flag indicating the (equivalent) rotation direction of the source-detector, with 1 representing clockwise and -1 representing counterclockwise. The paired images $f_{\theta_1}(\vec{x})$ and $f_{\theta_{1+T/2}}(\vec{x})$ are registered with $j = 1$, increasing sequentially.

2) Apply the proposed GCC-PHAT (VW) image registration method to obtain the strengthened CCS for each paired, rotated LCT image. Then, estimate the horizontal and vertical indices (i.e., pixel offsets $(Np_x, Np_y)$) corresponding to the highest peak (maximum value) from each CCS, yielding $T/2$ sets of pixel offsets, $\langle\{Np_{x_j}\}_{j=1}^{T/2}, \{Np_{y_j}\}_{j=1}^{T/2}\rangle$.

3) Remove it when there is an outlier in each set of estimated pixel offsets, $\{Np_{x_j}\}_{j=1}^{T/2}$ or $\{Np_{y_j}\}_{j=1}^{T/2}$, then select the optimized pixel offset results, $\langle\widehat{Np_x}, \widehat{Np_y}\rangle$, according to **Criterion 1**.

4) Utilize Eqs. (8) and (10) to calculate geometric errors $\langle\widehat{\delta s}, \widehat{\delta l}\rangle$ from the final estimated pixel offsets $\langle\widehat{Np_x}, \widehat{Np_y}\rangle$.

5) Substitute the final estimated errors $\langle\widehat{\delta s}, \widehat{\delta l}\rangle$ into the new projection geometry structure (i.e., yielding new calibration vectors) when programming the reconstruction process using the Astra toolbox [39].

Furthermore, Algo. 1 gives the pseudo code of our geometric error estimation method for correction.

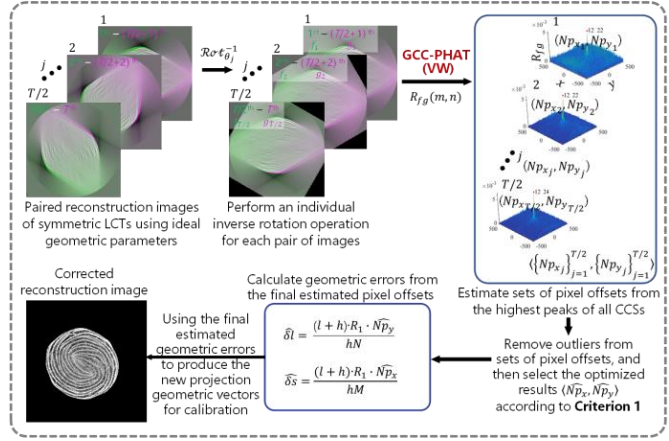

Fig. 7. Framework of the propsed geometric artifact correction method for SMLCT.

## IV. EXPERIMENTS AND GENERALIZATIONS

### A. Simulated experiments

To evaluate the performance of the proposed method, we set some known geometric errors to numerically scan a 3D Forbild phantom with 512 ($M$) × 512 ($N$) × 150 ($N_z$) voxels (as the ground-truth and cutting off up-down parts considering cone-beam artifacts, see Fig. 8(a)), getting the central 2D fan-beam projection data and directly reconstructing all LCT images. From these 2D LCT images, our method was adopted to estimate some sensitive geometric errors and obtain calibrated images, and then we utilized RMSE, SSIM, and profiles of some details to assess calibration results. Note that no other comparative methods were introduced in this experiment, as the applicable SMLCT method proposed by Yu *et al.* [13] is very primitive, inefficient, and subjective.

---

**Algorithm 1** Geometric error estimation method for correction

---

**Require:** $f_{\theta_i}(m,n)$, $i \in [1,T]$

1: Initialize $\widehat{\delta l}, \widehat{\delta s}, \{Np_{x_j}\}_{j=1}^{T/2}, \{Np_{y_j}\}_{j=1}^{T/2}$

2: **for** $j = 1 \to T/2$ **do**

3: $\langle f(m,n), g(m,n) \rangle \leftarrow \mathcal{R}ot_{\theta_j}^{-1}\left(\langle f_{\theta_j}(m,n), f_{\theta_{j+T/2}}(m,n)\rangle\right)$,
$\theta_j = (j-1) \cdot \mathcal{R}_{dir} \cdot \Delta\theta$

4: $G_{fg}(u,v) \leftarrow \mathcal{DFT}(f(m,n)) \cdot \overline{\mathcal{DFT}(g(m,n))}$;

5: $R_{fg}(m,n) \leftarrow \mathcal{DFT}^{-1}\left\{\frac{G_{fg}(u,v)}{|G_{fg}(u,v)|} \cdot B(u,v)\right\}$

6: $(Np_{x_j}, Np_{y_j}) \leftarrow \underset{m,n}{\mathrm{argmax}}(R_{fg}(m,n))$

7: **end for**

8: Remove outliers from $\langle\{Np_{x_j}\}_{j=1}^{T/2}, \{Np_{y_j}\}_{j=1}^{T/2}\rangle$, and then select the optimized results $\langle\widehat{Np_x}, \widehat{Np_y}\rangle$ according to **Criterion 1**

9: $\widehat{\delta l} \leftarrow \left((l+h) \cdot R_1 \cdot \widehat{Np_y}\right)/hN$

10: $\widehat{\delta s} \leftarrow \left((l+h) \cdot R_1 \cdot \widehat{Np_x}\right)/hM$

---

First, we only set two errors: $\overline{\delta l}$ = 2 mm and $\overline{\delta s}$ = 0.8 mm, to investigate the estimation accuracy of two sensitive errors in the 10-segment SMLCT geometry (i.e., $l$ = 15 mm, $h$ = 170 mm, $2s$ = 40 mm, number ($N_d \times N_h$) and size ($\Delta u$) of FPD units were 768 × 768 and 0.17 mm, $R_1$ = 11.88 mm, the FOV radius $r_1$ of RCT is only 4.99 mm, number $N_\lambda$ of projection views of LCT was 251, $T$ = 10. In noise-free data, GCC-PHAT (VW) estimated sets of pixel offsets include (33, -83) for the 1st - 6th, 2nd - 7th, and 4th - 9th LCTs, and (33, -84) for the 3rd - 8th and 5th - 10th LCTs, thus determined $(\widehat{Np_x}, \widehat{Np_y})$ = (33, -83) via **Criterion 1**, as shown in Fig. 8(b), and the final calculated geometric errors are about $\widehat{\delta l}$ = 2.096 mm and $\widehat{\delta s}$ = 0.833 mm. In noisy data (Poisson noise with $5 \times 10^3$ photon numbers exposed to each FPD pixel), Figure 8(c) respectively displays the estimated sets of pixel offsets, including (33, -82) for the 1st - 6th and 2nd - 7th LCTs, (33, -84) for the 3rd - 8th LCTs, and (33, -83) for the rest two sets of LCTs, calculating the mean $(\widehat{Np_x}, \widehat{Np_y})$ = (33, -82.8) according to **Criterion 1** to get the final estimated errors $\widehat{\delta l}$ = 2.091 mm and $\widehat{\delta s}$ = 0.833 mm for calibration. These estimated results reveal that the accuracy of our method is quite high, and the final corrected reconstruction images are impressive.

Next, we adjusted some geometric parameters to change into the other six-segment SMLCT geometry ($l$ = 13.75 mm, $h$ = 90.5 mm, $2s$ = 60 mm, $R_1$ = 12.87 mm, $r_1$ = 7.30 mm, $T$ = 6) and fused the four kinds of errors, including $\overline{\delta l}$ = 1.4 mm, $\overline{\delta h}$ = 1.5 mm, $\overline{\delta s}$ = 0.2 mm, and $\overline{\delta u}$ = 5 FPD pixels, to further verify our method's effectiveness. When there is no noise or even noise (with the same noise conditions as above), our method estimates the pixel offsets of all symmetric LCT reconstruction images as (12, -41), that is, the calculated sensitive geometric errors are $\widehat{\delta l}$ = 1.187 mm and $\widehat{\delta s}$ = 0.347 mm, the final corrected image quality has significantly improved compared to before calibration, as listed in Fig. 8(d).

Basing on the geometry of the first 10-segment SMLCT experiment and continuous to fuse new four kinds of errors: $\overline{\delta l}$ = -1.2 mm, $\overline{\delta h}$ = 2.5 mm, $\overline{\delta s}$ = 0.8 mm, and $\overline{\delta u}$ = -7 FPD pixels, as exhibited in Fig. 8(e), our method estimated the pixel offsets of all symmetric LCTs are (27, 53) with noise-free data, and it also got sets of pixel offsets with noisy data, including (27, 52) for the 1st - 6th and 2nd - 7th LCTs, and (27, 53) for the other three sets of symmetric LCTs. In these two conditions, we can both determine an optimal pixel offset (27, 53) and obtain the required $\widehat{\delta l}$ = -1.338 mm, $\widehat{\delta s}$ = 0.682 mm, and a clear corrected image.

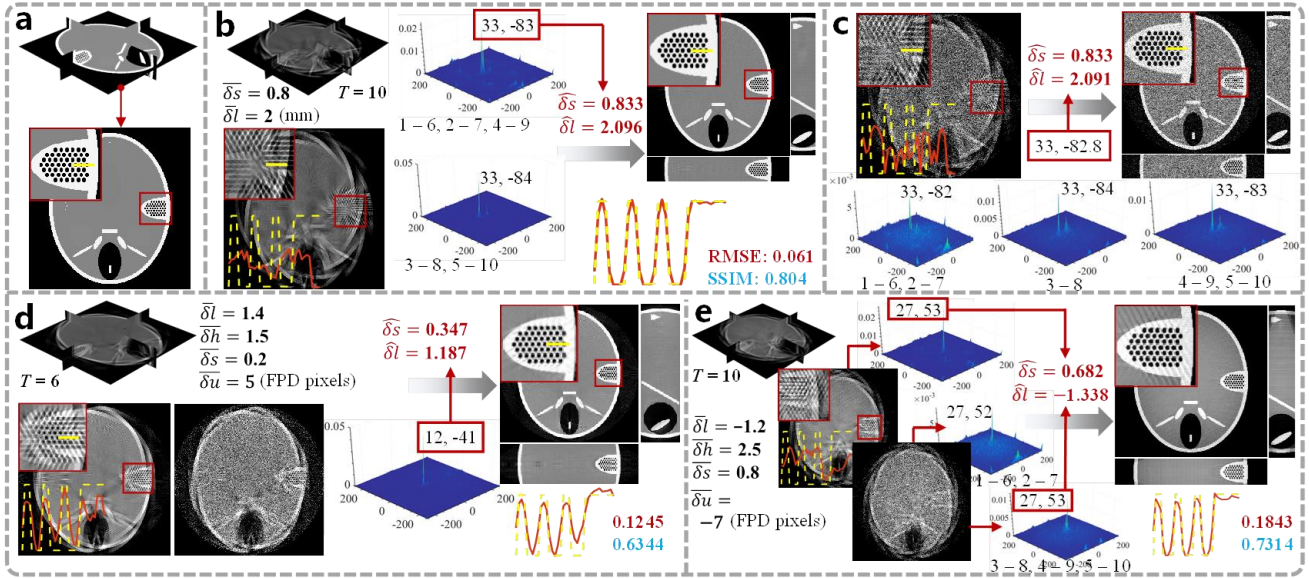

Fig. 8. Experimental results for verifying the proposed correction method on the simulated data. (a) Forbild phantom with being cut off for simulated scanning; (b) Uncorrected and our method corrected reconstruction results with only presetting $\overline{\delta l}$ and $\overline{\delta s}$, $T$ = 10, and noise-free data; (c) Previous uncorrected and final corrected results after adding Poisson noise to the simulated projection based on (b); (d) and (e) both display uncorrected and corrected immediate images and results, with four different types and magnitudes of geometric errors preset. Both figures include scenarios with noise-free and noisy projection data, respectively, under conditions of $T$ = 6 and $T$ = 10. Previous uncorrected and final noise-free corrected results display the profiles at the local high frequency details and the metrics of RMSE (red) and SSIM (blue) located at the lower right corner.

### B. Real experiments

To verify the effectiveness of our proposed calibration method for real SMLCT, we both performed real experiments for two equivalent SMLCT implementation **Modes 1** and **2**. First, we used the real bud projection data set [9] scanned in **Mode 1** to conduct a geometric self-calibration experiment, the scanning parameters include: $l$ = 13.75 mm, $h$ = 106.5 mm, $2s$ = 35 mm, $N_\lambda$ = 400, $N_d \times N_h$ = 1024 × 1024, $\Delta u$ = 0.127 mm, $R_1$ = 6.66 mm, reconstructed matrix: $M \times N \times N_z$ = 1024 × 1024 × 1024, $T$ = 6. If the above ideal geometric parameters were used to reconstruct an volume image, as shown in Fig. 8(a), severe artifacts hindered the observation of internal details. For the 1st - 4th and 2nd - 5th LCTs, the estimated pixel offsets both are $(\widehat{Np_x}, \widehat{Np_y})$ = (12, 22), whereas (12, 24) for the 3rd - 6th LCTs (three CCS figures can be found in Fig. 7), so the determined (12, 22) were substituted into Eqs. (8) and (10) to get $\widehat{\delta l}$ = -0.162 mm, $\widehat{\delta s}$ = 0.088 mm. The reconstructed bud volume via our method has a high definition for some micro-structures, some slices and rendered volume are displayed in Fig. 8(b) and (c), respectively.

**Table 1** Pixel offset estimation results based on GCC-PHAT (VW) image registration for real SMLCT-scanned data.

| Tea balls | 1 (1st-8th LCTs) | 2 (2nd-9th LCTs) | 3 (3rd-10th LCTs) | 4 (4th-11th LCTs) | 5 (5th-12th LCTs) | 6 (6th-13th LCTs) | 7 (7th-14th LCTs) | |
|---|---|---|---|---|---|---|---|---|
| $(Np_x,$ in CCS | 140, 72 | 140, 71 | 139, 70 | 140, 72 | 140, 70 | 140, 74 | 140, 70 | $\widehat{Np_x}$ = 140, $\widehat{Np_y}$ = 71.286 |
| **Batteries** | | | | | | | | |
| $(Np_x,$ in CCS | 26, 43 | 26, 43 | 26, 43 | 26, 43 | 26, 44 | 26, 43 | 26, 43 | $\widehat{Np_x}$ = 26, $\widehat{Np_y}$ = 43 |

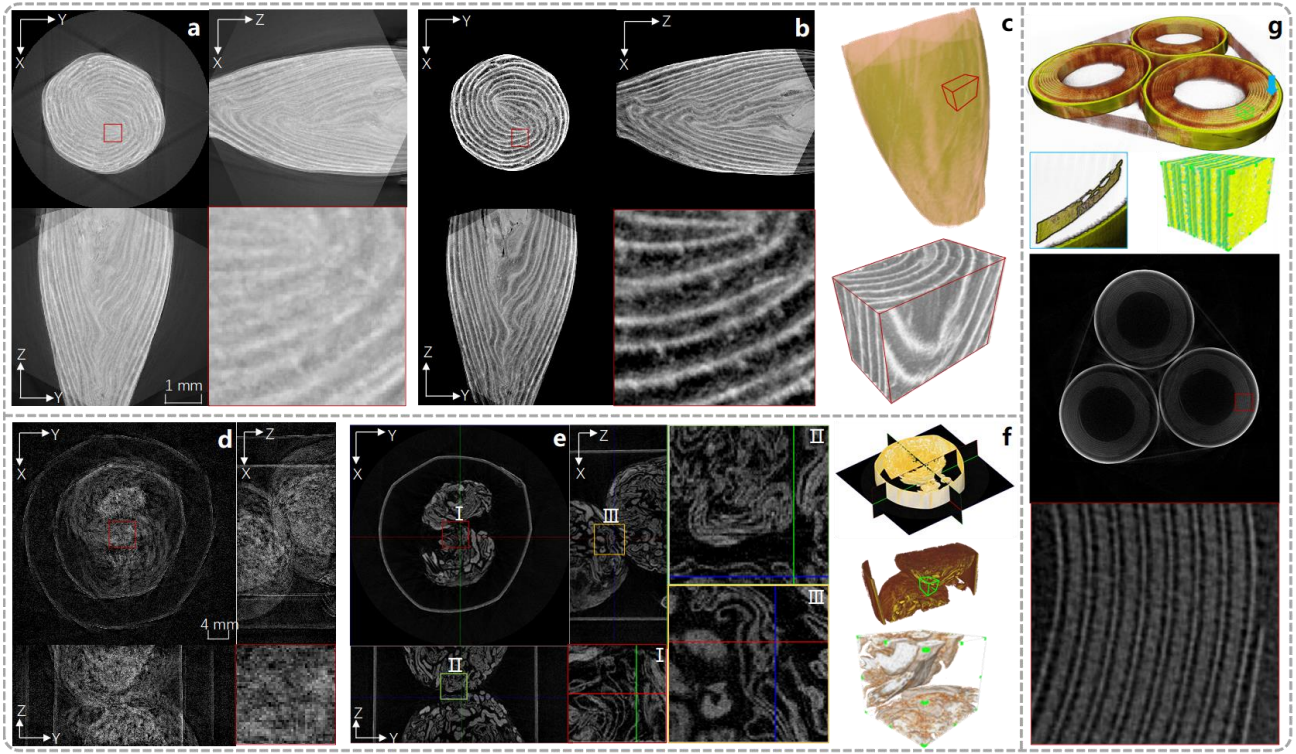

Fig. 9. Real experiment results of our correction method. (a) and (b) are non-corrected and corrected real results, including central orthogonal slices and detail views, in the reconstructed flower bud volume, respectively; (c) Rendering bud volume and partial enlarged view through our correction; (d) and (e) are non-corrected and corrected real results of the tea balls, and (e) shows three detailed views in three orthogonal slices: I, II, and III; (f) Rendering total volume and partial enlarged volume of tea balls; (g) Corrected reconstruction results of lithium batteries, including rendering total and partial volumes and grayscale figures.

Then, we used the equipment shown in Fig. 1(c) to complete SMLCT scanning of **Mode 2** for a cylinder containing several tea balls. Some parameters set were $l$ = 34 mm, $h$ = 280 mm, $2s$ = 66 mm, $N_\lambda$ = 400, $N_d \times N_h$ = 1536 × 1536, $\Delta u$ = 0.085 mm, $R_1$ = 21.34 mm ($r_1$ = 6.92 mm), $M \times N \times N_z$ = 1024 × 1024 × 420, $T$ = 14. Continously, with the same parameters, we conducted our correction method for scanned lithium batteries displayed in Fig. 1. All calculated CCS figures through GCC-PHAT (VW) image registration and the final estimated pixel offsets are shown in Table 1. For the final estimated pixel offsets, $(\widehat{Np_x}, \widehat{Np_y})$ = (140, 71.286) and (26, 43), in the tea balls and three batteries, we obtain the corresponding values ($\widehat{\delta l}$ = -1.666 mm, $\widehat{\delta s}$ = 3.272 mm) and ($\widehat{\delta l}$ = -1.005 mm, $\widehat{\delta s}$ = 0.608 mm) for geometric artifact correction, respectively. Compared to the non-corrected reconstructed volume of the tea balls (see central slices in Fig. 9(d)), Figure 9(e) demonstrates that our estimated correction significantly clarifies the final volume, with almost no visible geometric artifacts, as shown in the three orthogonal central slices and their locally enlarged details (see Figs. I, II, and III in Fig. 9(e)). Figure 9(f) displays the overall and local rendering volumes of the corrected tea balls. Regarding the previously non-corrected lithium batteries shown in Fig. 1(g), our final corrected results are presented as global and local magnified rendering volumes, along with global and local 2D grayscale slices, as shown in Fig. 9(g). In summary, these real experimental results fully and convincingly confirmed the impressive and practical performance of our proposed correction method.

### C. Generalized experiments for other CT scanning modes

Though our work aims at a special case in the MLCT scanning mode—SMLCT, for the actual asymmetric LCT geometry, we can also control the equipment to acquire two segment symmeric LCT projections and then use our method to correct geometric artifacts. Moreover, this correction work derived from the special SMLCT geometry is not lacking in generality and can be further extended for other CT scanning modes, e.g., the common RCT, including full- and half-scan, and even segmented helical CT (SHCT) generalized from SMLCT to extend FOV in the bi-directions [47,48].

Without loss of generality, for the common full- or half-scan RCT scanning modes, a local arc scanning trajectory is also a limited-angle CT scan, sharing some similar theories and characteristics to the MLCT we summarized earlier. Even for the half-scan RCT, there are two symmetric arc trajectries (e.g., two-segment angle range of $[-\gamma_m, \gamma_m]$ and $[\pi - \gamma_m, \pi + \gamma_m]$, where $\gamma_m$ is the maximum half fan-beam-angle), as shown in Fig. 10(a). Therefore, the proposed method can be applied to correct artifacts related to the most sensitive parameter—COR (which allows us to equivalently estimate and correct the detector's central offset $\widehat{\delta u}$). To verify this viewpoint, we continuously set some simulated parameters ($l$ =

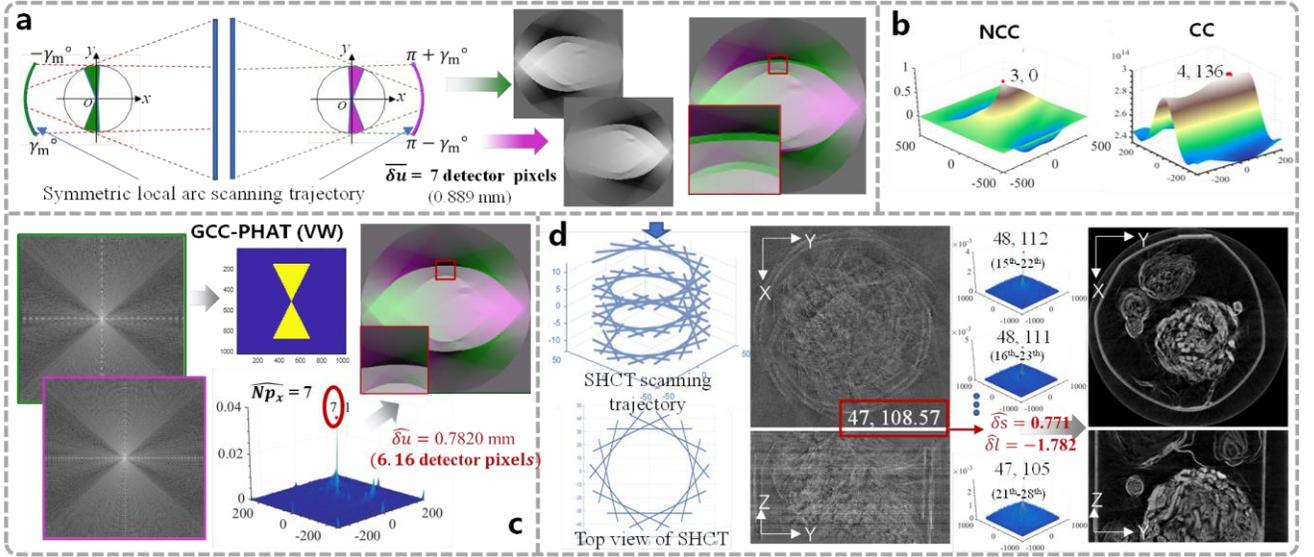

Fig. 10. Generalized experiments of our proposed method. (a) shows two-segment symmetric arc scanning trajectories and their fan-type Radon regions, individual reconstruction images, and a non-corrected overlaying image; (b) CCS maps of NCC and CC for two symmetric geometric reconstruction images; (c) depicts the individual frequency spetrum of two images, an applied bow-tie filter and CCS map, and a final corrected and overlaying image; (d) shows the SHCT scanning trajectory and uncorrected reconstruction slices, CCS maps of all symmetric slant-line trajectories in the middle turn, and the final corrected slices.

13.75 mm, $h$ = 106.5 mm, $N_d \times N_h$ = 1024 × 1024, $\Delta u$ = 0.127 mm, $r_1$ = 6.540 mm, $M \times N$ = 512 × 512) and a COR error (throught set $\overline{\delta u}$ = 7 pixels, i.e., the detector was offset by 7 pixel units in the $u$-axis direction). Figure 10(b) displays some wrong registration results, whereas the estimated number of pixel offsets $\widehat{Np_x}$ = 7 throught our GCC-PHAT (VW) (see the frequency spectrum diagrams of two-segment images, applied bow-tie filter, CCS map, and registration result in Fig. 10(c)), from CCS was obtained by registering reconstructed images with two symmetric angular ranges containing redundant parts, and then we substituted it into the following new relationship between $Np_x$ and $\delta u$ (rather than $\delta s$),

$$\delta u = \frac{(l+h) \cdot r_1 \cdot Np_x}{lM}. \tag{14}$$

Here, we estimated $\widehat{\delta u}$ = 0.7820 mm (approximately 6.16 detector pixels), which is very close to the ground truth, with a bias of less than one detector pixel.

Additionally, the SHCT scanning mode is an extension of SMLCT, with a top view equivalent to that of SMLCT. Our research group previously developed approximate volume reconstruction algorithms for SHCT, which produce acceptable quality when, and only when, the SHCT pitch is small—similar to the standard helical scan. As a result, the geometric artifacts in SHCT with small pitches have a distribution similar to those in SMLCT. Therefore, our correction method can also be applied to SHCT. We performed the real SHCT scanning for tea balls, and some key parameters were set as: $l$ = 34 mm, $h$ = 280 mm, the slant-line trajectory length 64 mm, number of turns $N_r$ = 3, $N_\lambda$ = 300, $N_d \times N_h$ = 1536 × 1536, $\Delta u$ = 0.085 mm, $R_1$ = 20.49 mm ($r_1$ = 6.92 mm), $M \times N \times N_z$ = 1400 × 1400 × 800, $T$ = 14, and the slant angle of the slant-line trajectory 2.106°, and the axial movement distance of the object after each diagonal scan 0.571 mm (some more details can be found in Ref. [48]). Figure 10(d) illustrates the scanning trajectory of SHCT, along with two reconstruction slices showing uncorrected and nearly invisible structures, CCS maps of our GCC-PHAT (VW), and the final corrected reconstruction slices. We used the projection data of a middle turn in SHCT to reconstruct 2D central cross-section images for correction. These estimated pixel offsets from the symmetric 15th - 22th, 16th -23th, ... , 21th -28th segments were (48, 112), (48, 111), (47, 108), (46, 108), (47, 111), (47, 105), and (47, 105), respectively, so the final pixel offsets $(\widehat{Np_x}, \widehat{Np_y})$ = (47, 108.57), the geometric errors ($\widehat{\delta l}$ = -1.782 mm, $\widehat{\delta s}$ = 0.771 mm), and the corrected clear reconstruction results are depicted in Fig. 10(d).

## V. DISCUSSION

### A. Discusssion on our geometric artifact correction

Note that our work is not designed to accurately calculate the precise geometric errors but rather to eliminate geometric artifacts as much as possible for the SMLCT scanning mode. Therefore, for the four geometric parameters that influence iso-center offset and cause severe geometric artifacts, we selected only two equivalent parameters for correction. As seen in the simulated experiment results of Fig. 8, our approach seems not to achieve absolutely perfect correction, but it significantly simplifies the process, and the reconstruction quality after correction is still highly satisfactory for practical engineering applications. In practical applications, it is preferred to simplify the number of correction steps and parameters. Any cumbersome or redundant operations are likely to hinder widespread application. Moreover, this simplification will hardly affect the actual resolution measurement, as our method ensures that the magnification remains accurate; otherwise, windmill-shaped artifacts could occur.

*B. Discusssion on our GCC-PHAT limited-angle image registration*

Various types of image registration methods have been developed, including those based on feature points, mutual information, histogram matching, regions, deep learning, and so on [49–51]. While some registration methods can match complex and challenging images in certain scenarios, they often involve complex computations and are difficult to implement. In contrast, our symmetric LCT images require only rigid translation registration, with our primary focus on quickly obtaining pixel offsets after registration. For this purpose, mutual information-based methods are more suitable. The proposed GCC-PHAT (VW) image registration method, which falls under the mutual information category, is particularly well-suited for limited-angle CT images, including but not limited to those in LCT. In our investigation, the GCC-PHAT algorithm has not yet been introduced into the field of image registration, mainly because the effective frequency range of general images is usually unknown. It opens a new path for limited-angle CT image registration. In the GCC-PHAT (VW), by applying weighting only to the VFR rather than whitening the entire spectrum, we preserve the relative phase information of specific frequencies. After the inverse Fourier transform, this retained phase information generates larger correlation peaks at non-zero positions in the spatial domain. This approach effectively becomes a specialized form of phase correlation for image registration [52,53].

*C. Future works*

For the requirement of non-ultra large extended FOV, we can apply the short-scan mode, wherea these is not symmetrical LCTs. In this case, calculating geometric deviations using the offset is not straightforward. In the future, we will attempt to correct geometric artifacts by utilizing the collinear redundant data of adjacent segments and the conjugate redundant data of the first and last segments. We also continue to investigate more efficient methods for correcting more parameters.

## VI. CONCLUSION

In this paper, to address the practical and challenging geometric artifacts commonly encountered in the typical MLCT scanning mode—SMLCT for extended FOV reconstruction—we propose a highly efficient, non-phantom geometric artifact correction method. Our work primarily focuses on developing the relevant correction theories and methods while also exploring their broader applications, including extensions to other CT scanning modes. We analyze the formation mechanisms of geometric artifacts with distinct features arising from two sensitive geometric errors, summarize some key properties behind these, and establish mathematical relationships that link the pixel offsets of two symmetric LCT reconstruction images to these sensitive geometric parameters. To continuously and well accomplish this, we designed a GCC-PHAT-based image registration method specifically for this kind of limited-angle CT scan, which efficiently and accurately extracts pixel offsets for correction, outperforming some classical registration methods.

**Declaration of competing interest**

The authors declare that they have no conflict of interest.

**Data availability**

Data will be made available on request.

ACKNOWLEDGMENTS

This work was supported in part by the National Natural Science Foundation of China (Grant No.: 52075133), CGN-HIT Advanced Nuclear and New Energy Research Institute (Grant No.: CGN-HIT202215). The authors would like to thank Prof. Fenglin Liu and Dr. Haijun Yu for their valuable discussions and constructive suggestions, and also thank Ubicomap Engineer Chunmin Zhang for their assistance in the actual experimental implementation.